\documentclass{article}

\PassOptionsToPackage{numbers}{natbib}
\usepackage[final]{nips_2017}

\usepackage[utf8]{inputenc} 
\usepackage[T1]{fontenc}    
\usepackage{hyperref}       
\usepackage{url}            
\usepackage{booktabs}       
\usepackage{amsfonts}       
\usepackage{nicefrac}       
\usepackage{microtype}      
\usepackage{graphicx}
\usepackage{amsmath}
\usepackage[noend]{algpseudocode}
\usepackage{algorithm,caption}
\usepackage{xcolor}
\usepackage{subcaption}
\usepackage{multicol}
\usepackage{xr}

\title{Robust and Efficient Transfer Learning with Hidden Parameter Markov Decision Processes}

\author{
	Taylor Killian\thanks{Both contributed equally as primary authors} \\
	\texttt{taylorkillian@g.harvard.edu}\\
	Harvard University
\And
	Samuel Daulton$^{*}$ \\
	\texttt{sdaulton@g.harvard.edu}\\
	Harvard University, Facebook\thanks{Current affiliation, joined afterward}
\AND
	\qquad\quad George Konidaris \\
	\qquad\quad\texttt{gdk@cs.brown.edu} \\
	\qquad\quad Brown University
\And \qquad \quad Finale Doshi-Velez \\
	\qquad\quad\texttt{finale@seas.harvard.edu}\\
	\qquad\quad Harvard University
}

\begin{document}

\maketitle

\begin{abstract}
We introduce a new formulation of the Hidden Parameter Markov Decision
Process (HiP-MDP), a framework for modeling families of related tasks
using low-dimensional latent embeddings.  Our new framework correctly
models the joint uncertainty in the latent parameters and the state
space.  We also replace the original Gaussian Process-based model with
a Bayesian Neural Network, enabling more scalable inference.  Thus, we
expand the scope of the HiP-MDP to applications with higher dimensions
and more complex dynamics.
\end{abstract}

\section{Introduction}
\label{sec:intro}
The world is filled with families of tasks with similar, but not
identical, dynamics. For example, consider the task of training a
robot to swing a bat with unknown length $l$ and mass $m$.  The task
is a member of a family of bat-swinging tasks. If a robot has already
learned to swing several bats with various lengths and masses {\small
  $\{(l_i,m_i)\}_{i=1}^N$}, then the robot should learn to
swing a new bat with length $l'$ and mass $m'$ more efficiently than
learning from scratch. That is, it is grossly inefficient to develop a
control policy from scratch each time a unique task is encountered.

The Hidden Parameter Markov Decision Process (HiP-MDP)
\citep{doshivelez2016HiP_MDP} was developed to address this type of
transfer learning, where optimal policies are adapted to subtle
variations within tasks in an efficient and robust
manner. Specifically, the HiP-MDP paradigm introduced a
low-dimensional latent task parameterization $w_b$ that, combined with
a state and action, completely describes the system's dynamics
{\small $T(s'|s,a,w_b)$}. However, the original formulation did not
account for nonlinear interactions between the latent parameterization and the
state space when approximating these dynamics, which required all
states to be visited during training. In addition, the original
framework scaled poorly because it used Gaussian Processes (GPs) as 
basis functions for approximating the task's dynamics.

We present a new HiP-MDP formulation that models interactions
between the latent parameters $w_b$ and the state $s$ when
transitioning to state $s'$ after taking action $a$. We do so by
including the latent parameters $w_b$, the state $s$, and the action
$a$ as \emph{input} to a Bayesian Neural Network (BNN). The BNN both
learns the common transition dynamics for a family of tasks and models how the unique variations of a particular instance impact the instance's overall dynamics.
Embedding the latent parameters in this way allows for more accurate
uncertainty estimation and more robust transfer when learning a
control policy for a new and possibly unique task instance. Our
formulation also inherits several desirable properties of BNNs: it can model multimodal and
heteroskedastic transition functions, inference scales to data
large in both dimension and number of
samples, and all output dimensions are jointly modeled, which reduces computation and
increases predictive accuracy~\cite{depeweg2016learning}. Herein, a BNN can capture complex
dynamical systems with highly non-linear interactions between state
dimensions. Furthermore, model uncertainty is easily quantified
through the BNN's output variance.  Thus, we can scale to larger domains than previously possible.  

We use the improved HiP-MDP formulation to
develop control policies for acting in a simple two-dimensional
navigation domain, playing acrobot~\cite{sutton1998rl}, and designing
treatment plans for simulated patients with
HIV~\cite{ernst2006clinical}. The HiP-MDP rapidly determines the
dynamics of new instances, enabling us to quickly find near-optimal
instance-specific control policies.

\section{Background}
\label{sec:prev_work}

\paragraph{Model-based reinforcement learning}
We consider reinforcement learning (RL) problems in which an agent
acts in a continuous state space $S \subseteq \mathbb{R}^D$ and a
discrete action space $A$. We assume that the environment has some
true transition dynamics {\small $T(s' | s , a)$}, unknown to the
agent, and we are given a reward function {\small $R(s,a): S\times A
  \rightarrow \mathbb{R}$} that provides the utility of taking action
$a$ from state $s$. In the model-based reinforcement learning setting,
our goal is to learn an approximate transition function {\small
  $\hat{T}(s'|s,a)$} based on observed transitions {\small $(s , a,
  s')$} and then use $\hat{T}(s'|s,a)$ to learn a policy $a = \pi(s)$
that maximizes long-term expected rewards $E[\sum_t \gamma^t r_t]$,
where {\small$\gamma \in (0,1]$} governs the relative importance of 
immediate and future rewards.

\paragraph{HiP-MDPs} 
A HiP-MDP \citep{doshivelez2016HiP_MDP} describes a \textit{family} of Markov Decision Processes (MDPs) and is defined by the tuple
{\small$\left\{S , A , W , T , R , \gamma , P_{W}\right\}$}, where
{\small$S$} is the set of states $s$, {\small$A$} is the set of
actions $a$, and $R$ is the reward function. The transition dynamics
{\small$T\left(s' | s, a,w_b\right)$} for each task instance $b$
depend on the value of the hidden parameters {\small$w_b\in W$}; for
each instance, the parameters $w_b$ are drawn from prior {\small$P_W$}.
The HiP-MDP framework assumes that a finite-dimensional array of hidden
parameters $w_b$ can fully specify variations among the true task dynamics. It also
assumes the system dynamics are invariant during a task and the 
agent is signaled when one task ends and another begins.

\paragraph{Bayesian Neural Networks}
A Bayesian Neural Network (BNN) is a neural network, {\small
  $f(\cdot,\cdot;\mathcal{W})$}, in which the parameters {\small
  $\mathcal{W}$} are random variables with some prior {\small
  $P(\mathcal{W})$} \citep{mackay1992practical}. We
place independent Gaussian priors on each parameter {\small
  $P(\mathcal{W}) = \prod_{w \in \mathcal{W}} \mathcal{N}(w; \mu,
  \sigma^2)$}. Exact Bayesian inference for the posterior over
parameters {\small $P(\mathcal{W}|\{(s',s,a)\})$} is intractable, but
several recent techniques have been developed to scale inference in
BNNs
\citep{blundell2015weight,gal2016dropout,hernandez2016black,neal1992bayesian}.
As probabilistic models, BNNs reduce the tendency of neural networks
to overfit in the presence of low amounts of data---just as GPs do. 
In general, training a BNN is more computationally efficient than a GP~\citep{hernandez2016black}, 
while still providing coherent uncertainty measurements. Specifically, predictive 
distributions can be calculated by taking averages over samples of 
{\small $\mathcal{W}$} from an approximated posterior distribution over the parameters. As such,
BNNs are being adopted in the estimation of stochastic dynamical
systems~\citep{depeweg2016learning,gal2016improving}.\nocite{dietrich1997fast,quinonero2005unifying,snelson2006sparse}.

\section{A HiP-MDP with Joint-Uncertainty}
\label{sec:model}

The original HiP-MDP transition function models variation across task
instances as:\footnote{We
	present a simplified version that omits their filtering variables
	$z_{kad}\in\{0,1\}$ to make the parallels between our formulation and
	the original more explicit; our simplification does not change any key properties.}
\begin{align}
s'_d &\approx \sum_{k=1}^K w_{bk} \hat{T}^{(\textit{GP})}_{kad}(s)+\epsilon \nonumber \\
w_{bk} &\sim \mathcal{N}(\mu_{w_k},\sigma^2_w) \nonumber \\
\epsilon &\sim \mathcal{N}(0,\sigma^2_{ad}),
\label{eq:linear}
\end{align}
where $s_d$ is the $d^{th}$ dimension of $s$. Each basis transition
function {\small $\hat{T}_{kad}$} (indexed by the $k^{th}$ latent parameter, the action $a$, and the dimension $d$) 
is a GP using only $s$ as input, linearly combined with instance-specific weights
$w_{bk}$. Inference involves learning the parameters for the GP basis
functions and the weights for each instance. GPs can robustly approximate stochastic state transitions in continuous dynamical systems in
model-based reinforcement
learning~\citep{deisenroth2011pilco,rasmussen2003gpinrl,rasmussen2006gp4ml}. GPs
have also been widely used in transfer learning outside of
RL (e.g.~\citep{bonilla2008multitask}).

While this formulation is expressive, it has limitations.  The primary
limitation is that the uncertainty in the latent parameters $w_{kb}$
is modeled independently of the agent's state uncertainty.  Hence, the
model does not account for interactions between the latent
parameterization $w_b$ and the state $s$. As a result,
\citet{doshivelez2016HiP_MDP} required that each task instance
$b$ performed the \emph{same set} of state-action combinations $(s,a)$
during training. While such training may sometimes be
possible---e.g. robots that can be driven to identical positions---it
is onerous at best and \emph{impossible} for other systems such as
human patients. The secondary limitation is that each output
dimension $s_d$ is modeled separately as a collection of GP basis functions {\small
  $\{\hat{T}_{kad}\}_{k=1}^K$}. The basis functions for output
dimension $s_d$ are independent of the basis functions for output
dimension $s_{d'}$, for $d \neq d'$. Hence, the model does not account
for correlation between output dimensions. Modeling such
correlations typically requires knowledge of how dimensions interact 
in the approximated dynamical system~\citep{alvarez2012kernels,genton2015cross}.
We choose not to constrain the HiP-MDP with such a priori knowledge since
the aim is to provide basis functions that can ascertain these
relationships through observed transitions.

To overcome these limitations, we include the instance-specific
weights $w_b$ as input to the transition function and model all
dimensions of the output jointly:
\begin{align}
s' &\approx \hat{T}^{(\textit{BNN})}(s,a,w_b) + \epsilon \nonumber \\
w_b &\sim \mathcal{N}\left(\mu_w, \Sigma_b\right) \nonumber \\
\epsilon &\sim \mathcal{N}\left(0,\sigma^2_n\right).
\label{eq:joint}
\end{align}
This critical modeling change eliminates \emph{all} of the above
limitations: we can learn \emph{directly} from data as observed---which 
is abundant in many industrial and health domains---and no longer
require highly constrained training procedure. 
We can also capture the correlations in the outputs of these
domains, which occur in many natural processes.  

Finally, the computational demands of using GPs as the transition function limited the application of the original HiP-MDP
formulation to relatively small domains. In the following, we use a
BNN rather than a GP to model this transition function.
The computational requirements needed to learn
a GP-based transition function makes a direct comparison to our new BNN-based formulation infeasible within our experiments (Section~\ref{sec:experiments}). 
We demonstrate, in Appendix~\ref{apdx:gp2bnn}, that the BNN-based transition model \emph{far} exceeds the GP-based transition model in both computational and predictive performance. In addition, BNNs naturally
produce multi-dimensional outputs $s'$ without requiring prior 
knowledge of the relationships between dimensions. This
allows us to directly model output correlations between the $D$ state dimensions, leading to a more unified and coherent transition
model. Inference in a larger input space $s,a,w_b$ with a large number of samples is tractable using efficient approaches that let us---given a distribution {\small $P(\mathcal{W})$} and input-output
tuples $(s,a,s')$---estimate a distribution over the latent embedding
{\small $P(w_b)$}. This enables more robust, scalable transfer.   

\begin{figure}[btp]
	\centering
	\includegraphics[width=0.65\linewidth]{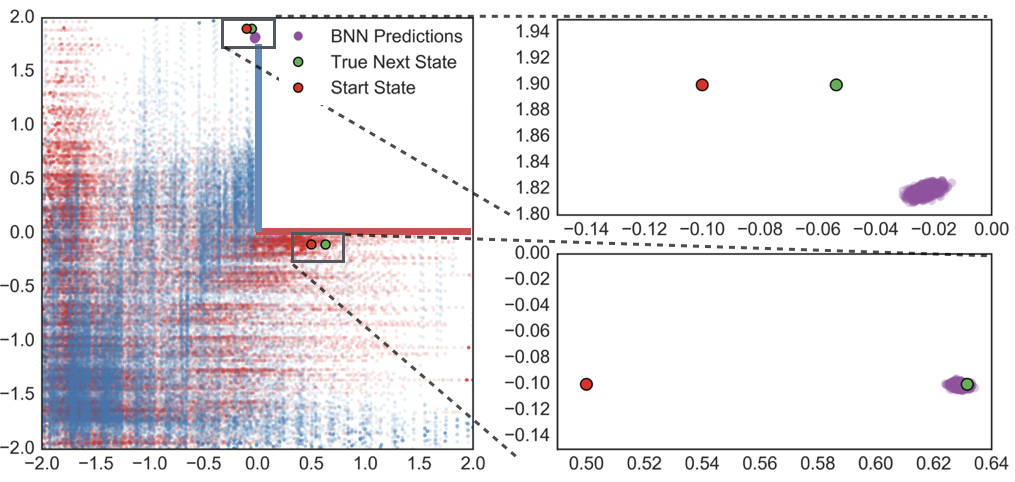} \caption{A demonstration of the HiPMDP modeling the joint uncertainty between the latent parameters $w_b$ and the state space. On the left, blue and red dots show the exploration during the red ($\theta_b=0$) and blue ($\theta_b=1$) instances. The latent parameters learned from the red instance are used predict transitions for taking action $E$ from an area of the state space either unexplored (top right) or explored (bottom right) during the red instance. The prediction variance provides an estimate of the joint uncertainty between the latent parameters $w_b$ and the state.}  \label{fig:toy_jointunc}
\end{figure}

\paragraph{Demonstration} 
We present a toy domain (Figure~\ref{fig:toy_jointunc}) where an agent
is tasked with navigating to a goal region. The state space is
continuous $(s\in(-2,2)^2)$, and action space is discrete
$(a\in\{N,E,S,W\})$. Task instances vary the following the domain aspects: the location of a wall that blocks access to the goal region (either to
the left of or below the goal region), the orientation of the cardinal
directions (i.e. whether taking action North moves the agent up or
down), and the direction of a nonlinear wind effect that increases as
the agent moves away from the start region. Ignoring the wall and grid
boundaries, the transition dynamics are:
\begin{align*}
\Delta x &= (-1)^{\theta_b}c\big(a_x - (1-\theta_b)\beta\sqrt{(x+1.5)^2+ (y+1.5)^2}\big)\\
\Delta y &= (-1)^{\theta_b}c\big(a_y - \theta_b\beta\sqrt{(x+1.5)^2+ (y+1.5)^2}\big)\\
a_x &=
\begin{cases}
1 & a \in \{E,W\}\\
0 & \text{otherwise}
\end{cases}\\
a_y&=
\begin{cases}
1 & a \in \{N,S\}\\
0 & \text{otherwise,}
\end{cases}
\end{align*}
where $c$ is the step-size (without wind), $\theta_b \in \{0,1\}$ indicates which of
the two classes the instance belongs to and $\beta \in (0,1)$
controls the influence of the wind and is fixed for all instances.
The agent is penalized for trying to cross a wall, and each step
incurs a small cost until the agent reaches the goal region, encouraging the agent to discover the goal region with the shortest route possible. An
episode terminates once the agent enters the goal region or after 100
time steps.

A linear function of the state $s$ and latent parameters $w_b$ would struggle to model 
both classes of instances ($\theta_b=0$ and $\theta_b=1$) in this domain because the 
state transition resulting from taking an action $a$ is a nonlinear function with interactions between the state and hidden parameter $\theta_b$.

By contrast, our new HiP-MDP model allows nonlinear interactions
between state and the latent parameters $w_b$, as well as jointly
models their uncertainty.  In Figure~\ref{fig:toy_jointunc}, this
produces measurable differences in transition uncertainty in regions
where there are few related observed transitions, even if there are many observations from unrelated
instances. Here, the HiP-MDP is trained on two instances
from distinct classes (shown in blue ($\theta_b=1$) and red
($\theta_b=0$) on the left). We display the uncertainty of the
transition function, {\small $\hat{T}$}, using the latent parameters
$w_{\text{red}}$ inferred for a red instance in two regions of the
domain: 1) an area explored during red instances and 2) an area not
explored under red instances, but explored with blue instances. The
transition uncertainty {\small $\hat{T}$} is three times larger in the
region where red instances have not been---even if many blue instances
have been there---than in regions where red instances have commonly
explored, demonstrating that the latent parameters can have different
effects on the transition uncertainty in different states.

\section{Inference}
\label{sec:inference}

Algorithm~\ref{alg:inference} summarizes the inference procedure for learning a 
policy for a new task instance $b$, facilitated by a pre-trained BNN for that task, and is
similar in structure to prior work \citep{deisenroth2011pilco,gal2016improving}.
The procedure involves several parts. Specifically, at the start of a new instance $b$, 
we have a global replay buffer $\mathcal{D}$ of all observed transitions $(s,a,r,s')$ and 
a posterior over the weights $\mathcal{W}$ for our BNN transition function {\small $\hat{T}$} 
learned with data from $\mathcal{D}$. The first objective is to quickly determine the latent 
embedding, $w_b$, of the current instance's specific dynamical variation as transitions 
$(s,a,s')$ are observed from the current instance. Transitions from instance $b$ are 
stored in both the global replay buffer $\mathcal{D}$ and an instance-specific replay buffer 
$\mathcal{D}_b$. The second objective is to develop an optimal control policy
using the transition model {\small $\hat{T}$} and learned latent parameters
$w_b$. The transition model {\small $\hat{T}$} and latent embedding
$w_b$ are separately updated via mini-batch stochastic gradient descent (SGD)
using Adam~\citep{kingma2014adam}. Using {\small $\hat{T}$} for
planning increases our sample efficiency as we reduce interactions with the environment. We describe each of these
parts in more detail below.

\subsection{Updating embedding $w_b$ and BNN parameters $\mathcal{W}$}

For each new instance, a new latent weighting $w_b$ is sampled from
the prior $P_W$ {\small(Alg.~\ref{alg:inference}, step 2)}, in preparation 
of estimating unobserved dynamics introduced by $\theta_b$. 
Next, we observe transitions $(s, a, r, s')$ from the task instance for an initial exploratory episode
{\small(Alg.~\ref{alg:inference}, steps 7-10)}. 
Given that data, we optimize the latent parameters $w_b$ to minimize 
the $\alpha$-divergence of the posterior predictions of $\hat{T}(s,a,w_b | \mathcal{W})$ 
and the true state transitions $s'$ {\small (step 3 in TuneModel)}~\citep{hernandez2016black}. 
Here, the minimization occurs by adjusting the latent embedding $w_b$ while holding the
BNN parameters $\mathcal{W}$ fixed. After an initial update of the
$w_b$ for a newly encountered instance, the parameters
$\mathcal{W}$ of the BNN transition function {\small $\hat{T}$} are
optimized {\small (step 4 in TuneModel)}. As the BNN is trained on multiple instances of a task, we found that the only additional data needed to refine the 
BNN and latent $w_b$ for some new instance can be provided by an initial exploratory episode. 
Otherwise, additional data from subsequent episodes can be used to 
further improve the BNN and latent estimates {\small(Alg.~\ref{alg:inference}, steps 11-14)}. 

The mini-batches used for optimizing the latent $w_b$ and BNN network parameters 
$\mathcal{W}$ are sampled from $\mathcal{D}_b$ with squared error
prioritization~\citep{moore1993prioritized}. We found that switching
between small updates to the latent parameters and small updates to
the BNN parameters led to the best transfer performance. If either the BNN network or 
latent parameters are updated too aggressively (having a 
large learning rate or excessive number of training epochs), the BNN 
disregards the latent parameters or state inputs respectively. After completing 
an instance, the BNN parameters and the latent parameters are updated using 
samples from global replay buffer $\mathcal{D}$. Specific modeling details such 
as number of epochs, learning rates, etc. are described in Appendix~\ref{apdx:model_specs}.

\subsection{Updating policy $\hat{\pi}_b$}
We construct an $\varepsilon$-greedy policy to select actions based on
an approximate action-value function {\small $\hat{Q}(s,a)$}. We
model the action value function {\small $\hat{Q}(s,a)$} with a Double
Deep Q Network (DDQN)~\citep{hasselt2016deep,mnih2015human}. The DDQN involves
training two networks (parametrized by $\theta$ and $\theta^-$
respectively), a primary Q-network, which informs the policy, and a
target Q-network, which is a slowly annealed copy of the primary
network {\small (step 9 of SimEp)} providing greater stability when
updating the policy $\hat{\pi}_b$ .

With the updated transition function, {\small $\hat{T}$}, we
approximate the environment when developing a control policy
{\small (SimEp)}. We simulate batches of entire episodes of length
$N_t$ using the approximate dynamical model {\small $\hat{T}$},
storing each transition in a fictional experience replay buffer
$\mathcal{D}^f_b$ {\small (steps 2-6 in SimEp)}. The primary network
parameters $\theta$ are updated via SGD every $N_\pi$ time steps
{\small (step 8 in SimEp)} to minimize the temporal-difference error
between the primary network's and the target network's Q-values. The
mini-batches used in the update are sampled from the fictional
experience replay buffer $\mathcal{D}^f_b$, using TD-error-based
prioritization~\citep{schaul2015prioritized}.

\begin{algorithm}[t]
	\caption{Learning a control policy w/ the HiP-MDP} 
	\label{alg:inference}
	\small
	\begin{multicols}{2}
		\begin{algorithmic}[1]
			\Statex \textbf{Input:} Global replay buffer $\mathcal{D}$, BNN transition function $\hat{T}$, initial state $s^0_b$
			\Procedure{LearnPolicy}{ $\mathcal{D},\hat{T},s^0_b$}
			\State Draw new $w_b\sim P_W$ 
			\State Randomly init. policy $\hat{\pi}_b$ $\theta,\theta^-$
			\State Init. instance replay buffer $\mathcal{D}_b$
			\State Init. fictional replay buffer $\mathcal{D}^f_b$
			\For {$i=0$ to $N_e$ episodes} 
			\Repeat
			\State Take action $a \leftarrow \hat{\pi}_b(s)$
			\State Store $\mathcal{D},\mathcal{D}_b \leftarrow (s, a, r, s' ,w_b)$ 
			\Until  {episode is complete}
			\If {$i = 0$ OR $\hat{T}$ is innaccurate}
			\State $\mathcal{D}_b,\mathcal{W},w_b\leftarrow$
			\Call{TuneModel}{$\mathcal{D}_b,\mathcal{W},w_b$}
			
			\For {$j=0$ to $N_f-1$ episodes} 
			\State $\mathcal{D}^f_b,\hat{\pi}_b\leftarrow$ \Call{SimEp}{$\mathcal{D}^f_b,\hat{T},w_b,\hat{\pi}_b,s^0_b$}
			\EndFor
			\EndIf
			\State $\mathcal{D}^f_b,\hat{\pi}_b\leftarrow$
			\Call{SimEp}{$\mathcal{D}^f_b,\hat{T},w_b,\hat{\pi}_b,s^0_b$}
			\EndFor
			\EndProcedure
		\end{algorithmic}
		\columnbreak
		\begin{algorithmic}[1]
			\Function{SimEp}{$\mathcal{D}^f_b,\hat{T},w_b,\hat{\pi}_b,s^0_b$}
			\For {$t=0$ to $N_t$ time steps} 
			\State Take action $a \leftarrow \hat{\pi}_b(s)$
			\State Approx. $\hat{s}'\leftarrow\hat{T}(s,a,w_b)$
			\State Calc. reward $\hat{r}\leftarrow R(s,a,\hat{s}')$
			\State Store $\mathcal{D}^f_b \leftarrow (s, a, \hat{r}, \hat{s}')$
			\If {$\mod( t, N_{\pi} ) = 0$}
			\State Update $\hat{\pi}_b$ via $\theta$ from $\mathcal{D}^f_b$
			\State $\theta^- \leftarrow  \tau\theta+(1-\tau)\theta^-$
			\EndIf
			\EndFor
			\State \Return {$\mathcal{D}^f_b,\hat{\pi}_b$}
			\EndFunction
		\end{algorithmic}
		\begin{algorithmic}[1]
			\Function{TuneModel}{$\mathcal{D}_b,\mathcal{W},w_b$}
			\For {$k=0$ to $N_u$ updates} 
			\State Update $w_b$ from $\mathcal{D}_b$
			\State Update $\mathcal{W}$ from $\mathcal{D}_b$
			\EndFor
			\State \Return {$\mathcal{D}_b,\mathcal{W},w_b$}
			\EndFunction
		\end{algorithmic}
	\end{multicols}
\end{algorithm}

\section{Experiments and Results}
\label{sec:experiments}

Now, we demonstrate the performance of the HiP-MDP with embedded 
latent parameters in transferring learning across various instances of the same task. 
We revisit the 2D demonstration problem from Section~\ref{sec:model}, as well as 
describe results on both the acrobot~\citep{sutton1998rl} and a more complex healthcare 
domain: prescribing effective HIV treatments~\citep{ernst2006clinical} to patients with varying physiologies.\footnote{Example code for training and evaluating a HiP-MDP, including the simulators used in this section, can be found at \ \texttt{http://github.com/dtak/hip-mdp-public}.}

For each of these domains, we compare our formulation of the HiP-MDP
with embedded latent parameters (equation~\ref{eq:joint}) with four
baselines (one model-free and three model-based) to demonstrate the efficiency of learning a policy for a new
instance $b$ using the HiP-MDP. These comparisons are made across the first handful of episodes encountered in a new task instance to highlight the advantage provided by transferring information through the HiP-MDP.  The `linear' baseline uses a BNN to
learn a set of basis functions that are linearly combined with the
parameters $w_b$ (used to approximate the
approach of \citet{doshivelez2016HiP_MDP}, equation~\ref{eq:linear}), which does not allow
interactions between states and weights. The `model-based from scratch'
baseline considers each task instance $b$ as unique; requiring the BNN
transition function to be trained only on observations made from the
current task instance. The `average' model baseline is constructed
under the assumption that a single transition function can be used for
every instance of the task; {\small $\hat{T}$} is trained from
observations of all task instances together. For all
model-based approaches, we replicated the HiP-MDP procedure as closely
as possible. The BNN was trained on observations from a single episode
before being used to generate a large batch of approximate transition
data, from which a policy is learned. Finally, the model-free baseline
learns a DDQN-policy directly from observations of the current
instance. 

For more information on the experimental specifications and long-run policy learning see Appendix~\ref{apdx:model_specs} and~\ref{apdx:ex_results}, respectively.

\subsection{Revisiting the 2D demonstration}

\begin{figure*}[h!]
	\begin{subfigure}[t]{0.495\textwidth}
		\centering	
		\includegraphics[width=0.625\linewidth]{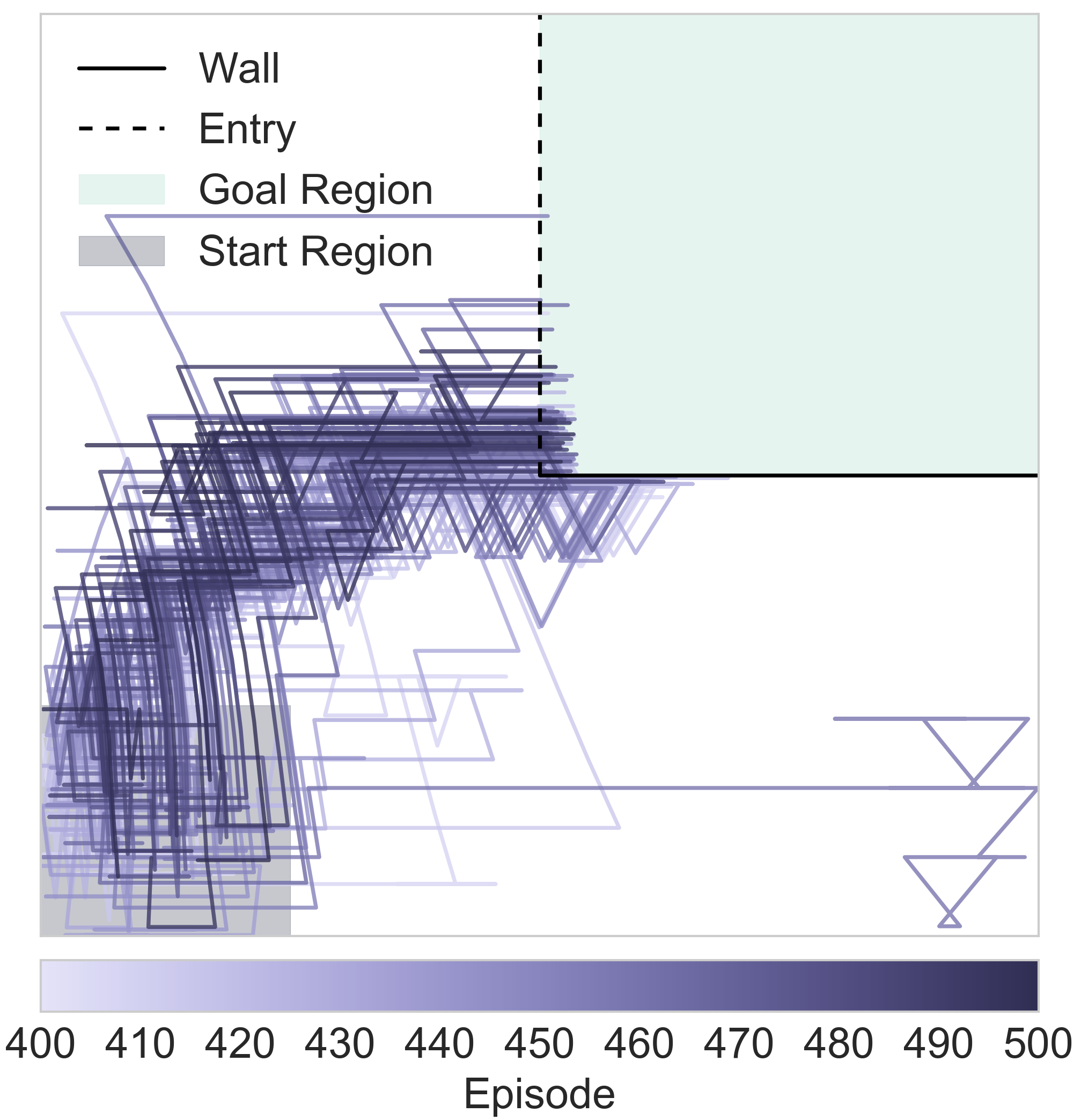}
		\caption{} \label{fig:grid_setup}
	\end{subfigure}%
	\begin{subfigure}[t]{0.495\textwidth}
		\centering	
		\includegraphics[width=0.9\linewidth]{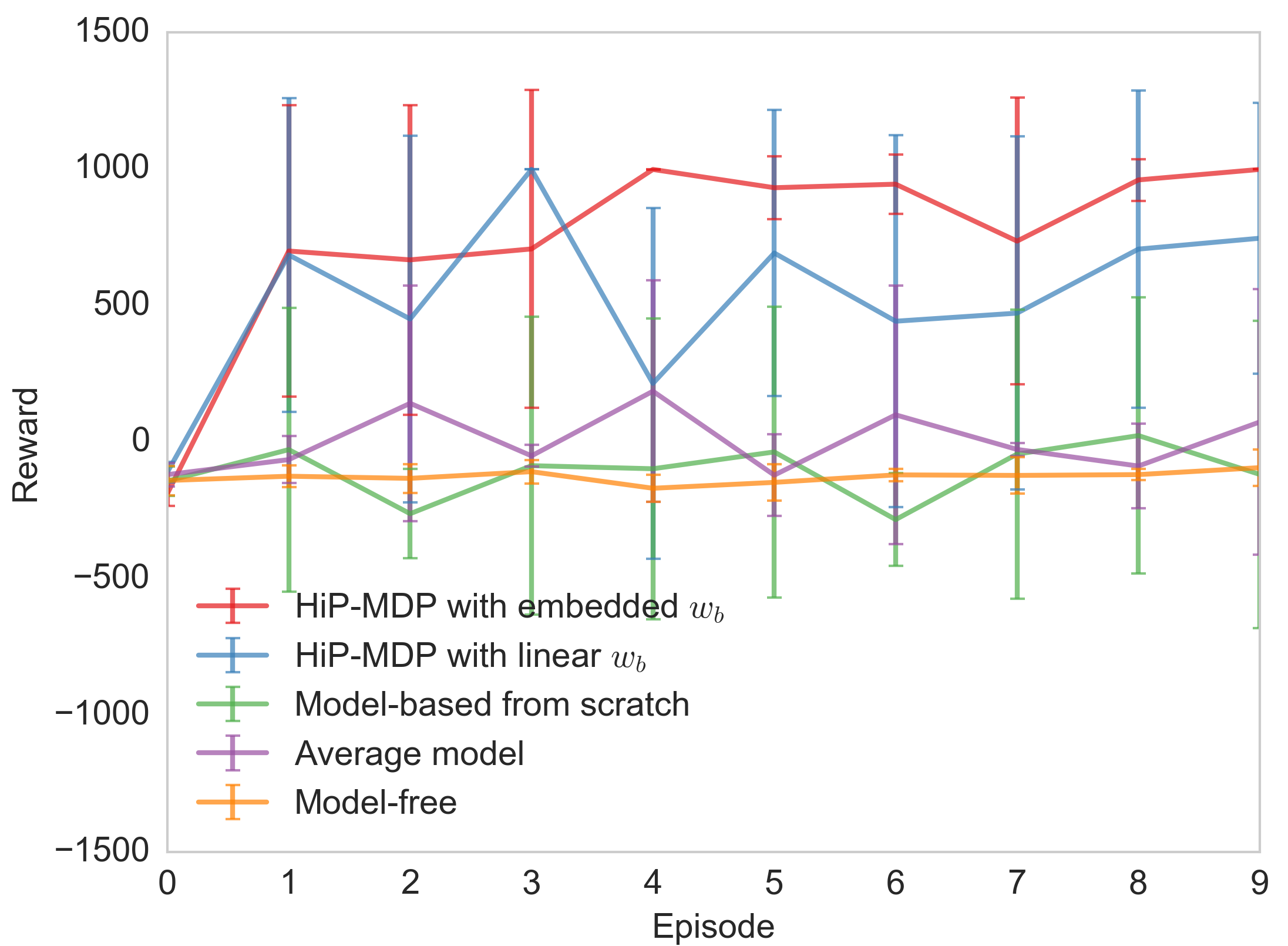}
		\caption{} \label{fig:grid_baselines}
	\end{subfigure}%
	\caption{(a) a demonstration of a model-free control policy, (b) a comparison of learning a policy at the outset of a new task instance $b$ using the HiP-MDP versus four benchmarks. The HiP-MDP with embedded $w_b$ outperforms all four benchmarks.}
        \label{fig:grid_results}
\end{figure*}

The HiP-MDP and the average model were supplied a transition model {\small $\hat{T}$} 
trained on two previous instances, one from each class, before being updated according 
to the procedure outlined in Sec.~\ref{sec:inference} for a newly encountered instance. 
After the first exploratory episode, the HiP-MDP has sufficiently determined the latent 
embedding, evidenced in Figure~\ref{fig:grid_baselines} where the developed policy 
clearly outperforms all four benchmarks. This implies that the transition model 
{\small $\hat{T}$} adequately provides the accuracy needed to develop an optimal 
policy, aided by the learned latent parametrization. 

The HiP-MDP with linear $w_b$ also quickly adapts to the new instance and learns 
a good policy. However, the HiP-MDP with linear $w_b$ is unable to model the 
nonlinear interaction between the latent parameters and the state. Therefore the 
model is less accurate and learns a less consistent policy than the HiP-MDP with 
embedded $w_b$. (See Figure~\ref{fig:grid_erros} in Appendix~\ref{apdx:pred_perf})

With single episode of data, the model trained from scratch on the current instance 
is not accurate enough to learn a good policy. Training a BNN from scratch requires 
more observations of the true dynamics than are necessary for the HiP-MDP to learn 
the latent parameterization and achieve a high level of accuracy. The model-free 
approach eventually learns an optimal policy, but requires significantly more observations
 to do so, as represented in Figure~\ref{fig:grid_setup}. The model-free approach has no improvement in the first 10 episodes. The 
 poor performance of the average model approach indicates that a single model cannot 
 adequately represent the dynamics of the different task instances. Hence, learning a latent 
 representation of the dynamics specific to each instance is crucial.

\subsection{Acrobot}

\begin{figure*}[t]
    \centering
	\begin{subfigure}[b]{0.4\textwidth}
		\centering	
		\includegraphics[width=0.5\linewidth]{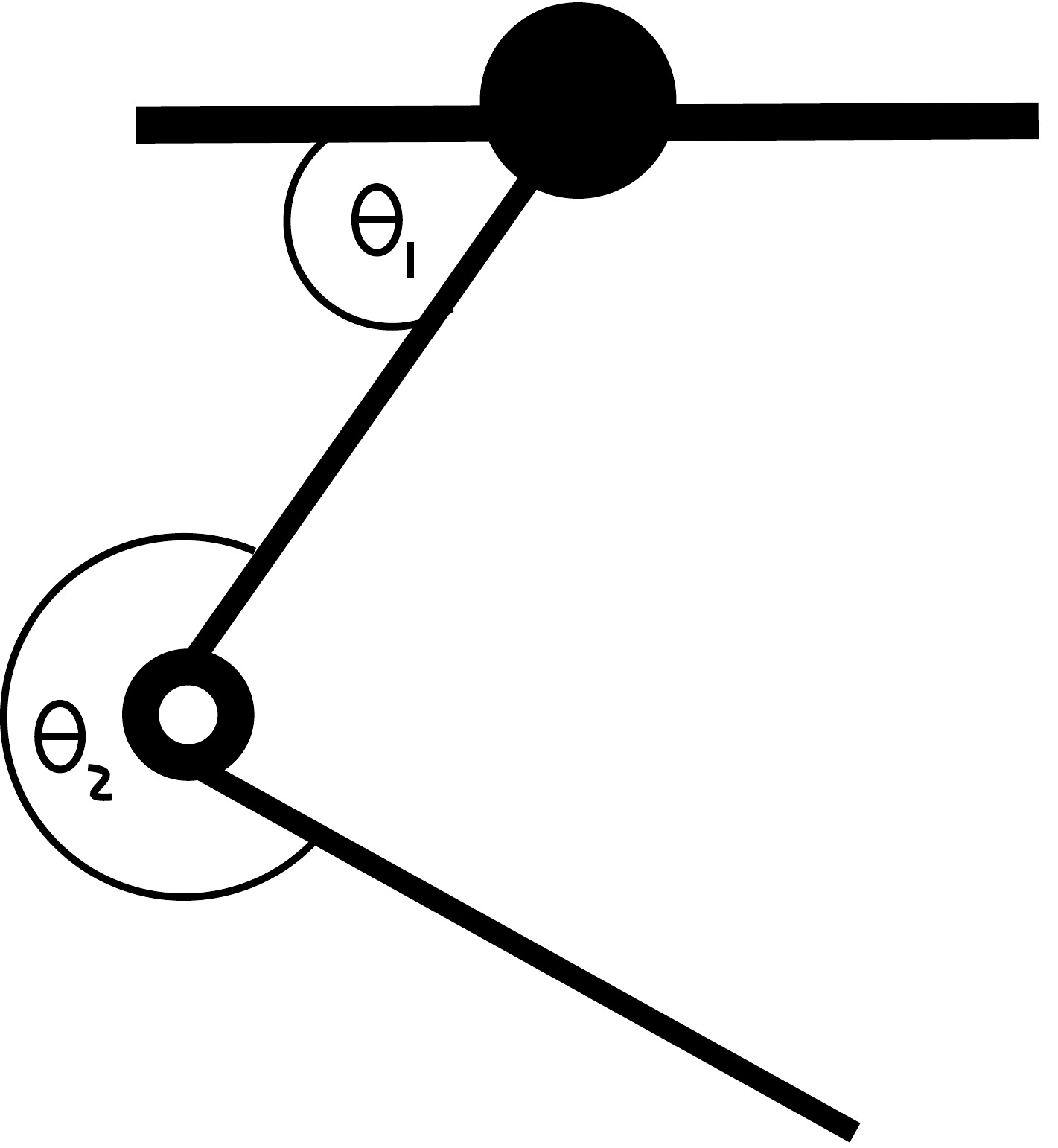}
		\caption{} \label{fig:acr_setup}
	\end{subfigure}%
	\begin{subfigure}[b]{0.45\textwidth}
		\centering	
		\includegraphics[width=\linewidth]{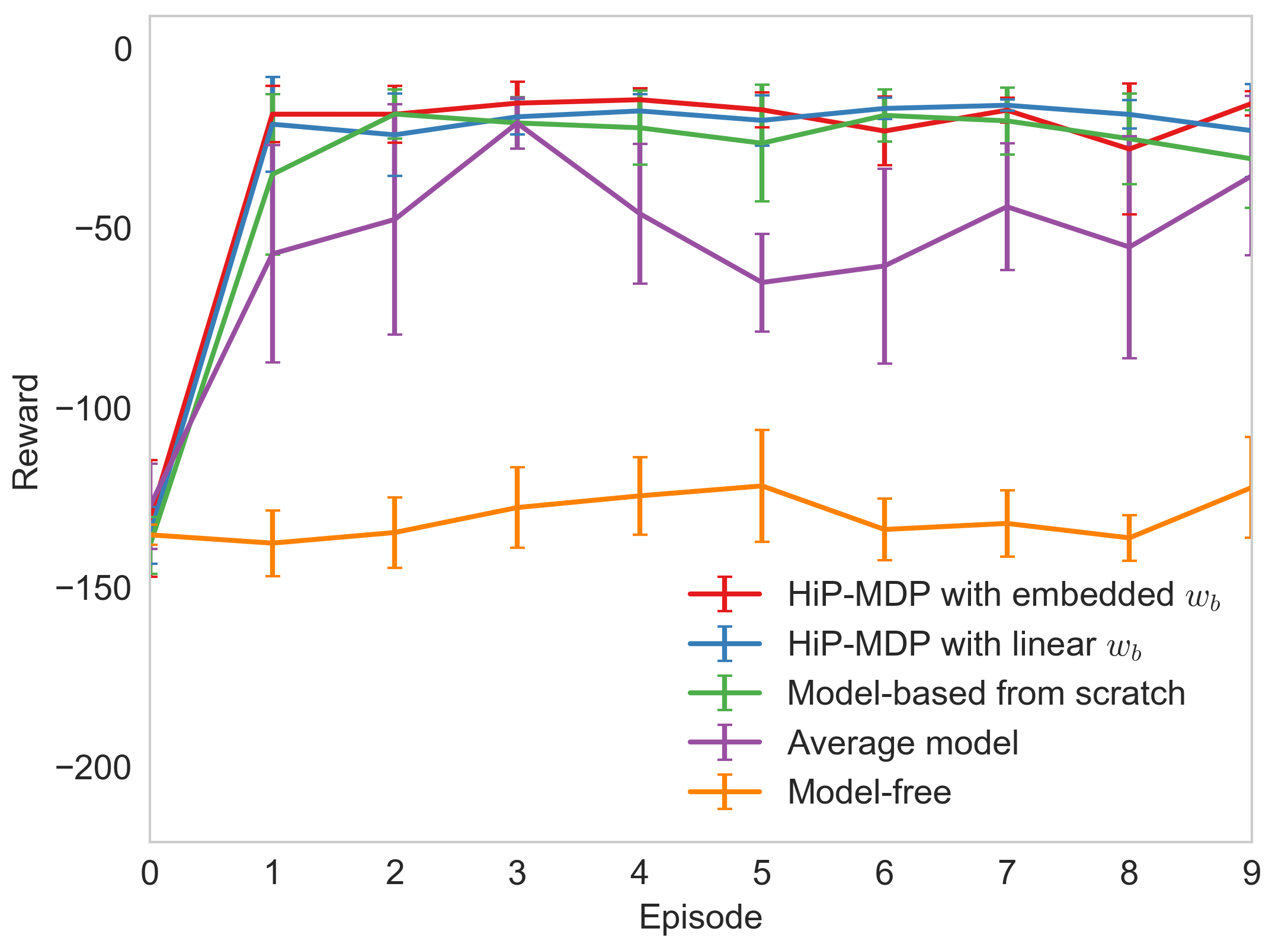}
		\caption{} \label{fig:acr_baselines}
	\end{subfigure}%
	\caption{(a) the acrobot domain, (b) a comparison of learning a policy for a new task instance $b$ using the HiP-MDP versus four benchmarks.}\label{fig:acr_results}
\end{figure*}

First introduced by~\citet{sutton1998rl}, acrobot is a canonical RL 
and control problem. The most common objective of this domain is for the agent 
to swing up a two-link pendulum by applying a positive, neutral, or negative torque 
on the joint between the two links {\small(see Figure~\ref{fig:acr_setup})}. These 
actions must be performed in sequence such that the tip of the bottom link reaches 
a predetermined height above the top of the pendulum. The state space consists of 
the angles $\theta_1$, $\theta_2$ and angular velocities $\dot\theta_1$, $\dot\theta_2$, 
with hidden parameters corresponding to the masses ($m_1$, $m_2$) and lengths 
($l_1$, $l_2$), of the two links.\footnote{The centers of mass and moments of inertia 
can also be varied. For our purposes we left them unperturbed.} See Appendix~\ref{apdx:dom_acrobot} 
for details on how these hidden parameters were varied to create different task instances. 
A policy learned on one setting of the acrobot will generally perform poorly on other 
settings of the system, as noted in {\small\cite{bai2013planning}}. Thus, subtle changes 
in the physical parameters require separate policies to adequately control the varied 
dynamical behavior introduced. This provides a perfect opportunity to apply the 
HiP-MDP to transfer between separate acrobot instances when learning a 
control policy $\hat{\pi}_b$ for the current instance.

Figure~\ref{fig:acr_baselines} shows that the HiP-MDP learns an optimal policy after a single episode, whereas all other model-based benchmarks required an additional episode of training. As in the toy example, the model-free approach eventually learns an optimal policy, but requires more time. 

\subsection{HIV treatment}

\begin{figure*}[t]
  \centering
	\begin{subfigure}[t]{0.4\textwidth}
		\centering
		\includegraphics[width=\linewidth]{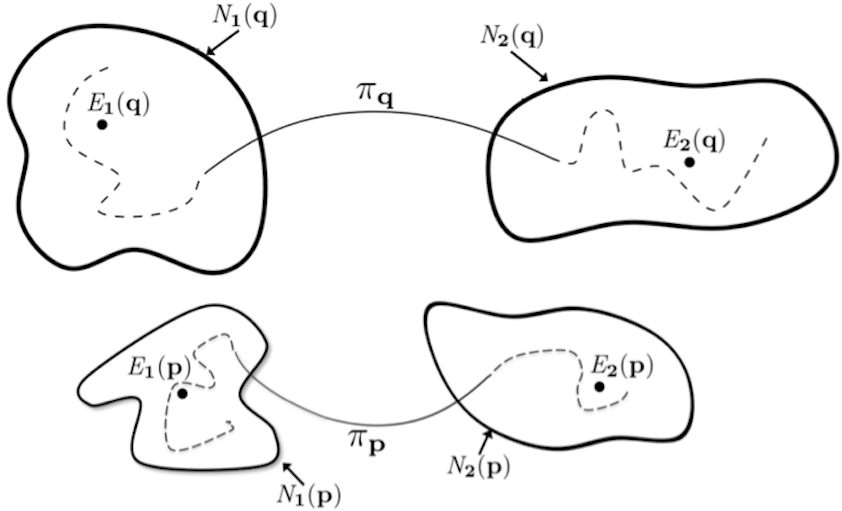}
		\caption{}\label{fig:hiv_setup}
	\end{subfigure}
	\begin{subfigure}[t]{0.45\textwidth}
		\centering	
		\includegraphics[width=\linewidth]{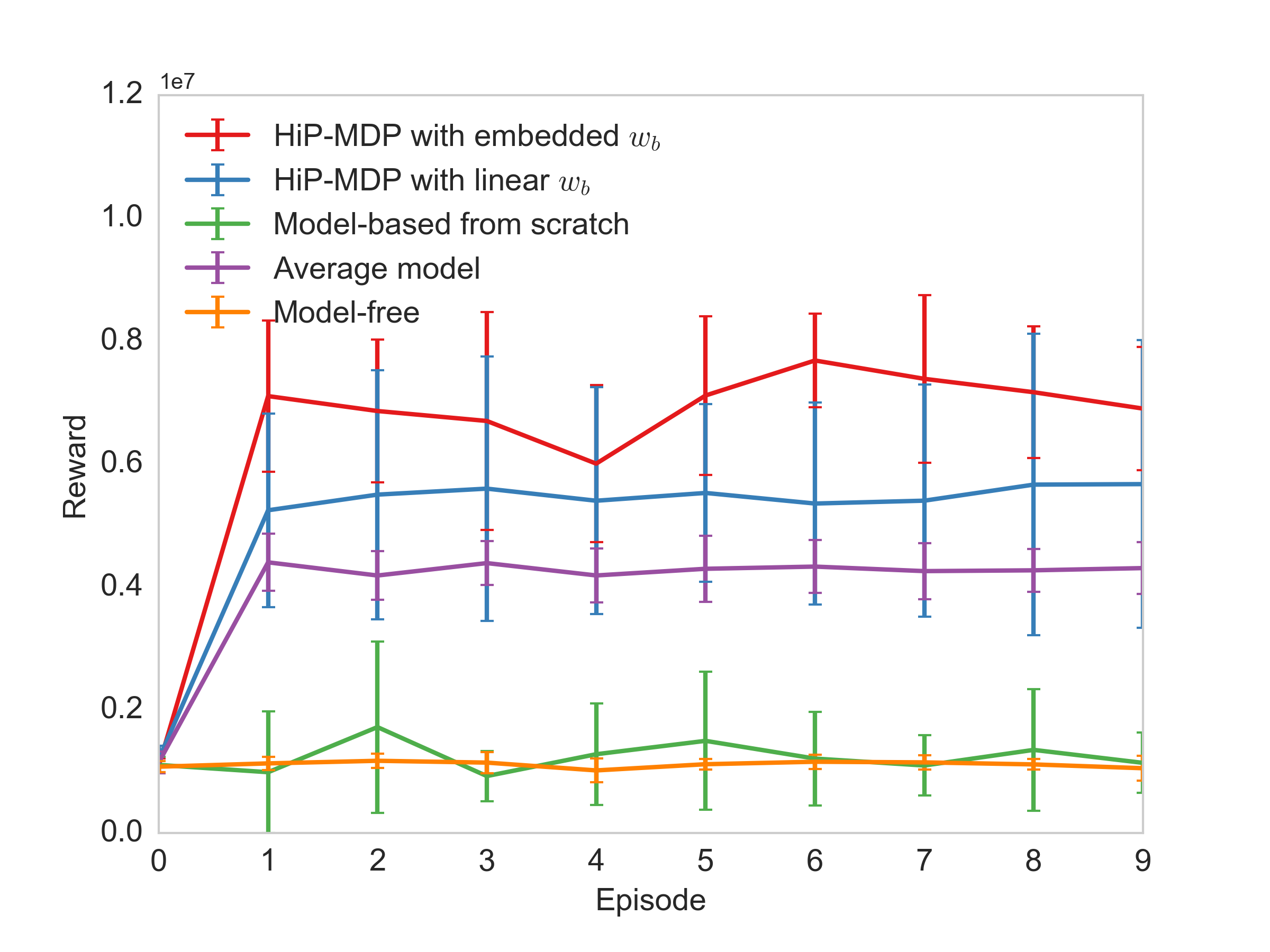}
		\caption{} \label{fig:hiv_baselines}
	\end{subfigure}%
	\caption{(a) a visual representation of a patient with HIV transitioning from an unhealthy steady state to a healthy steady state using a proper treatment schedule, (b) a comparison of learning a policy for a new task instance $b$ using the HiP-MDP versus four benchmarks.}\label{fig:hiv_results}
\end{figure*}

Determining effective treatment protocols for patients with HIV was
introduced as an RL problem by mathematically representing a
patient's physiological response to separate classes of
treatments~\citep{adams2004dynamic,ernst2006clinical}. In this model, the state of a patient's health is recorded
via 6 separate markers measured with a blood
test.\footnote{These markers are: the viral load ($V$), the number of
 healthy and infected CD4$^{+}$ T-lymphocytes ($T_1, \ T^{*}_1$,
 respectively), the number of healthy and infected macrophages ($T_2,
  \ T^{*}_2$, respectively), and the number of HIV-specific cytotoxic
 T-cells ($E$).} Patients are given one of four treatments on a
regular schedule. Either they are given treatment from one of two
classes of drugs, a mixture of the two treatments, or provided no
treatment (effectively a rest period). There are 22 hidden parameters in this
system that control a patient's specific physiology and dictate 
rates of virulence, cell birth, infection, and death. (See Appendix~\ref{apdx:dom_hiv} for more details.) 
The objective is to develop a treatment sequence that transitions
the patient from an unhealthy steady state to a healthy steady
state (Figure~\ref{fig:hiv_setup}, see \citet{adams2004dynamicfor 
a more thorough explanation)}. Small changes made to these 
parameters can greatly effect the behavior of
the system and therefore introduce separate steady state regions that
require unique policies to transition between them.

Figure~\ref{fig:hiv_baselines} shows that the HiP-MDP develops an optimal 
control policy after a single episode, learning an unmatched optimal policy in 
the shortest time. The HIV simulator is the most complex of our three 
domains, and the separation between each benchmark is more pronounced. 
Modeling a HIV dynamical system from scratch from a single 
episode of observations proved to be infeasible. The average model, which has been trained 
off a large batch of observations from related dynamical systems, learns a better 
policy. The HiP-MDP with linear $w_b$ is able to transfer knowledge from previous 
task instances and quickly learn the latent parameterization for this new instance, 
leading to an even better policy. However, the dynamical system contains nonlinear 
interactions between the latent parameters and the state space. Unlike the HiP-MDP 
with embedded $w_b$, the HiP-MDP with linear $w_b$ is unable to model those 
interactions. This demonstrates the superiority of the HiP-MDP with embedded 
$w_b$ for efficiently transferring knowledge between instances in highly complex domains.

\section{Related Work}

There has been a large body of work on solving single POMDP models
efficiently
\citep{brunskill2013sample,fern2010computational,kaelbling1998planning,rosman2016bayesian,williams2006scaling}.
In contrast, transfer learning approaches leverage training done on
one task to perform related tasks.  Strategies for transfer learning
include: latent variable models, reusing pre-trained model parameters,
and learning a mapping between separate tasks (see review in
\citep{taylor2009transfer}).

Our work falls into the latent variable model category.  Using latent
representation to relate tasks has been particularly popular in
robotics where similar physical movements can be exploited
across a variety of tasks and
platforms~\citep{delhaisse17transfer,gupta2017learning}.  In
\citet{chen2016pomdp}, these latent representations are encoded as
separate MDPs with an accompanying index that an agent learns while
adapting to observed variations in the
environment. \citet{bai2013planning} take a closely related approach
to our updated formulation of the HiP-MDP by incorporating estimates
of unknown or partially observed parameters of a known environmental
model and refining those estimates using model-based Bayesian RL. The
core difference between this and our work is that we learn the
transition model and the observed variations directly from the data
while \citet{bai2013planning} assume it is given and the specific
variations of the parameters are learned.  Also related are multi-task
approaches that train a single model for multiple tasks
simultaneously~\citep{bonilla2008multitask,caruana1998multitask}.
Finally, there have been many
applications of reinforcement learning
(e.g. \citep{moore2014reinforcement,shortreed2011informing,tenenbaum2010personalizing})
and transfer learning in the healthcare domain by identifying subgroups with similar response
(e.g. \citep{jaques2015multi,marivate2014quantifying,schulam2016integrative}).

More broadly, BNNs are powerful probabilistic inference models that allow for the estimation of stochastic dynamical systems~\citep{depeweg2016learning,gal2016improving}. Core to this functionality is their ability to represent both model uncertainty and transition stochasticity~\citep{kendall2017uncertainties}. Recent work decomposes these two forms of uncertainty to isolate the separate streams of information to improve learning. Our use of fixed latent variables as input to a BNN helps account for model uncertainty when transferring the pretrained BNN to a new instance of a task. Other approaches use stochastic latent variable inputs to introduce transition stochasticity~\citep{depeweg2017uncertainty,moerland2017learning}.

We view the HiP-MDP with latent embedding as a methodology that can facilitate personalization 
and do so robustly as it transfers knowledge of prior observations to the current 
instance. This approach can be especially useful in extending personalized care to groups of 
patients with similar diagnoses, but can also be extended to any control system where variations may be present.

\section{Discussion and Conclusion}

We present a new formulation for transfer learning among
related tasks with similar, but not identical dynamics, within the
HiP-MDP framework. Our approach leverages a latent
embedding---learned and optimized in an online fashion---to
approximate the true dynamics of a task. Our adjustment to the
HiP-MDP provides robust and efficient learning when faced with varied
dynamical systems, unique from those previously learned.  It is
able, by virtue of transfer learning, to rapidly determine optimal
control policies when faced with a unique instance.

The results in this work assume the presence of a large batch of
already-collected data.  This setting is common in many industrial and
health domains, where there may be months, sometimes years, worth of
operations data on plant function, product performance, or patient
health.  Even with large batches, each new instance still requires
collapsing the uncertainty around the instance-specific parameters in
order to quickly perform well on the task.  In
Section~\ref{sec:experiments}, we used a batch of transition data
from multiple instances of a task---without any artificial exploration
procedure---to train the BNN and learn the latent
parameterizations. Seeded with data from diverse task instances, the
BNN and latent parameters accounted for the variation between instances.

While we were primarily interested in settings where batches of
observational data exist, one might also be interested in more
traditional settings in which the first instance is completely new,
the second instance only has information from the first, etc.  In our
initial explorations, we found that one can indeed learn the BNN in 
an online manner for simpler domains.  However, even with
simple domains, the model-selection problem becomes more challenging:
an overly expressive BNN can overfit to the first few instances, and
have a hard time adapting when it sees data from an instance with very
different dynamics.  Model-selection approaches to allow the BNN to
learn online, starting from scratch, is an interesting future research
direction.

Another interesting extension is rapidly identifying the latent $w_b$.
Exploration to identify $w_b$ would supply the dynamical model with
the data from the regions of domain with the largest uncertainty. This
could lead to a more accurate latent representation of the observed
dynamics while also improving the overall accuracy of the transition
model. Also, we found training a DQN requires careful exploration
strategies. When exploration is constrained too early, the DQN quickly
converges to a suboptimal, deterministic policy––often choosing the
same action at each step. Training a DQN along the BNN's trajectories
of least certainty could lead to improved coverage of the domain and
result in more robust policies. The development of effective policies
would be greatly accelerated if exploration were more robust and
stable. One could also use the hidden parameters $w_b$ to learn a
policy directly.

Recognizing structure, through latent embeddings, between task
variations enables a form of transfer learning that is both robust and
efficient.  Our extension of the HiP-MDP demonstrates how embedding a
low-dimensional latent representation with the input of an approximate
dynamical model facilitates transfer and results in a more accurate
model of a complex dynamical system, as interactions between the
input state and the latent representation are modeled naturally. We
also model correlations in the output dimensions by replacing the GP
basis functions of the original HiP-MDP formulation with a BNN. The
BNN transition function scales \emph{significantly} better to larger and more complex problems. Our improvements to
the HiP-MDP provide a foundation for robust and efficient transfer
learning. Future improvements to this work will contribute to a
general transfer learning framework capable of addressing the most
nuanced and complex control problems.

\paragraph{Acknowledgements} We thank Mike Hughes, Andrew Miller, Jessica Forde, and Andrew Ross for their helpful conversations. TWK was supported by the MIT Lincoln Laboratory Lincoln Scholars Program. GDK is supported in part by the NIH R01MH109177. The content of this work is solely the responsibility of the authors and does not necessarily represent the official views of the NIH.

\small
\bibliographystyle{plainnat}
\bibliography{bibliography}
\normalsize
\appendix
\section*{Appendix}
\section{BNN-based transition functions with embedded latent weighting}
\label{apdx:gp2bnn}

\subsection{Scalability of BNN vs. GP based transition approximation} 
In this section, we demonstrate the computational motivation to replace the GP basis functions of the original HiP-MDP model~\citep{doshivelez2016HiP_MDP} with a single, stand-alone BNN as discussed in Sec.~\ref{sec:model}) using the 2D navigation domain. To fully motivate the replacement, we altered the GP-based model to accept the latent parameters $w_b$ and a one-hot encoded action as additional inputs to the transition model. This was done to investigate how the performance of the GP would scale with a higher input dimension; the original formulation of the HiP-MDP~\citep{doshivelez2016HiP_MDP} uses 2 input dimensions our proposed reformulation of the HiP-MDP uses 11 (2 from the state, 4 from the action and 5 from latent parameterization).

\begin{figure}[h!]
	\centering
	\includegraphics[width=0.7\linewidth]{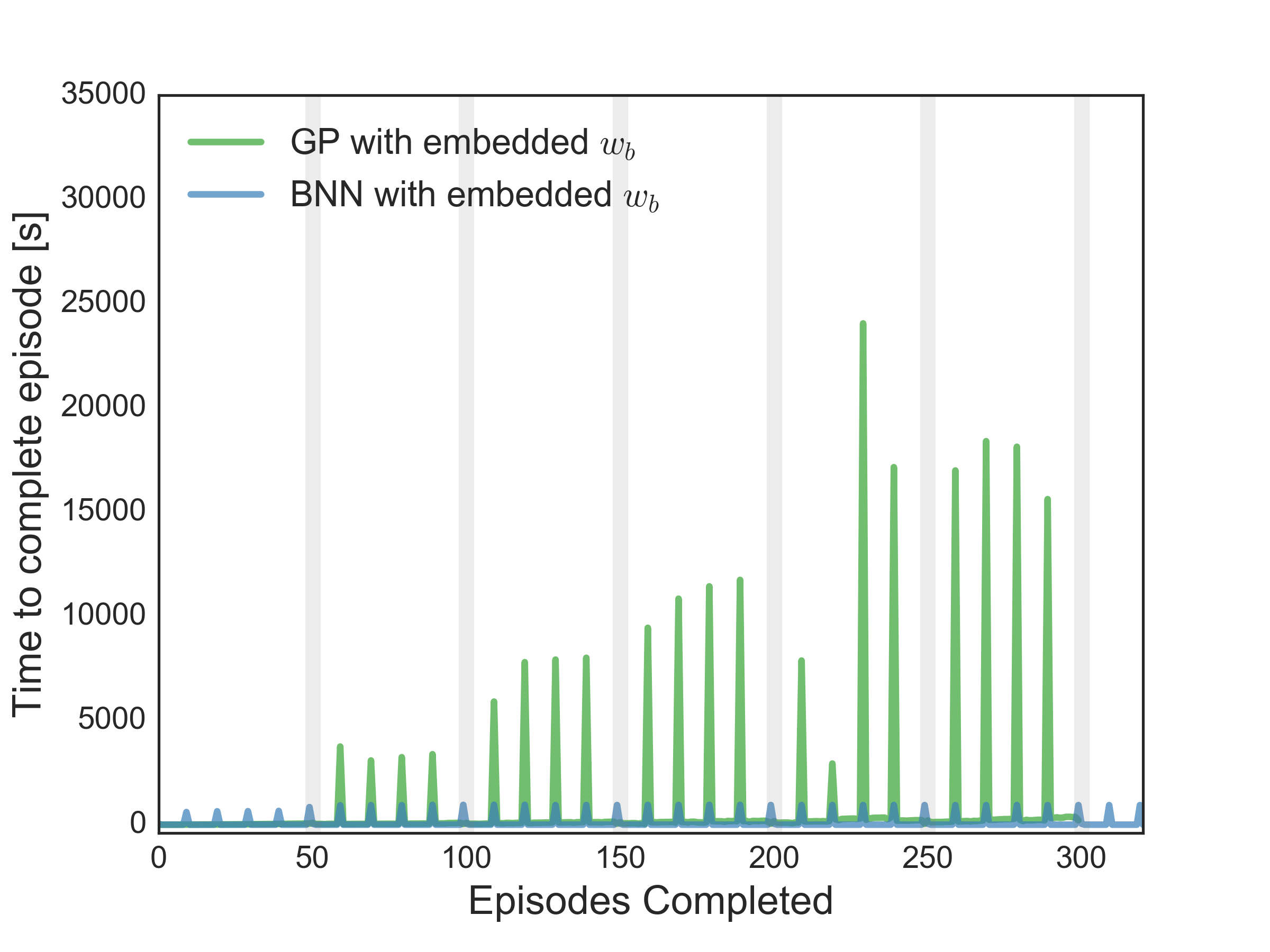}
	\caption{Here we show that the time to run an episode where the approximated transition model {\small $\hat{T}$} and latent parameters $w_b$ are updated every 10 episodes. In the 2D navigation domain, the completion time is relatively constant for the BNN, whereas the GP's completion time drastically increases as more data is collected to construct the transition model. }
	\label{fig:scalability}
\end{figure}

We directly compare the run-time performance of training both the GP-based and BNN-based HiP-MDP over 6 unique instances of the toy domain, with 50 episodes per instance.  Figure~\ref{fig:scalability} shows the running times (in seconds) for each episode of the GP-based HiP-MDP and the BNN-based HiP-MDP, with the transition model {\small $\hat{T}$} and latent parameters $w_b$ being updated after every 10 episodes.  In stark contrast with the increase in computation for the GP-based HiP-MDP, the BNN-based HiP-MDP has no increase in computation time as more data and further training is encountered. Training the BNN over the course of 300 separate episodes in the 2D toy domain was completed in a little more than 8 hours. In contrast, the GP-based HiP-MDP, trained on the 2D toy domain, took close to 70 hours to complete training on the same number of episodes. 

This significant increase in computation time using the GP-based HiP-MDP, on a relatively simple domain, prevented us from performing comparisons to the GP model on our other domains.  (We do realize that there is a large literature on making GPs more computationally efficient~\citep{dietrich1997fast,quinonero2005unifying,snelson2006sparse}; we chose to use BNNs because if we were going to make many inference approximations, it seemed reasonable to turn to a model that can easily capture heteroskedastic, multi-modal noise and correlated outputs.)

\subsection{Prediction performance: benefit of embedding the latent parameters} 
\label{apdx:pred_perf}
In the previous section, we justified replacing the GP basis functions in the HiP-MDP in favor of a BNN. In this section, we investigate the prediction performance of various BNN models to determine whether the latent embedding provides the desired effect of more robust and efficient transfer. The BNN models we will characterize here are those presented in Sec.~\ref{sec:experiments} used as baseline comparisons to the HiP-MDP with embedded latent parameters. These models are:

\begin{itemize}
\item HiP-MDP with embedded $w_b$
\item HiP-MDP with linear $w_b$
\item BNN learned from scratch, without any latent characterization of the dynamics
\item Average model: BNN trained on all data without any latent characterization of the dynamics.
\end{itemize}

\begin{figure*}[h!]
	\begin{subfigure}[t]{0.495\textwidth}
		\centering	
		\includegraphics[width=\linewidth]{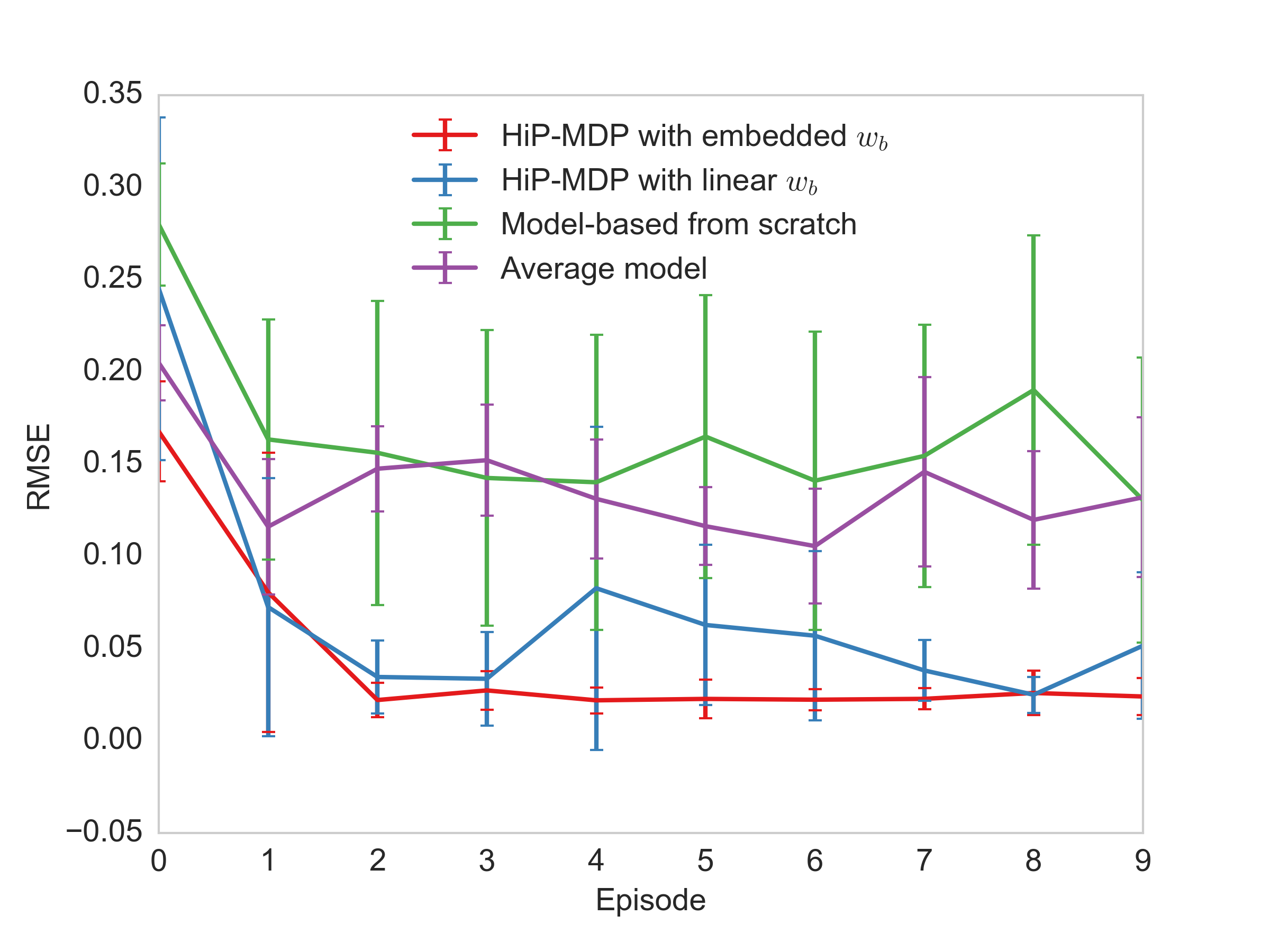}
		\caption{2D Toy Domain  $\left[\mathcal{S}\in\mathbb{R}^2\right]$} \label{fig:grid_erros}
	\end{subfigure}%
	\begin{subfigure}[t]{0.495\textwidth}
		\centering	
		\includegraphics[width=\linewidth]{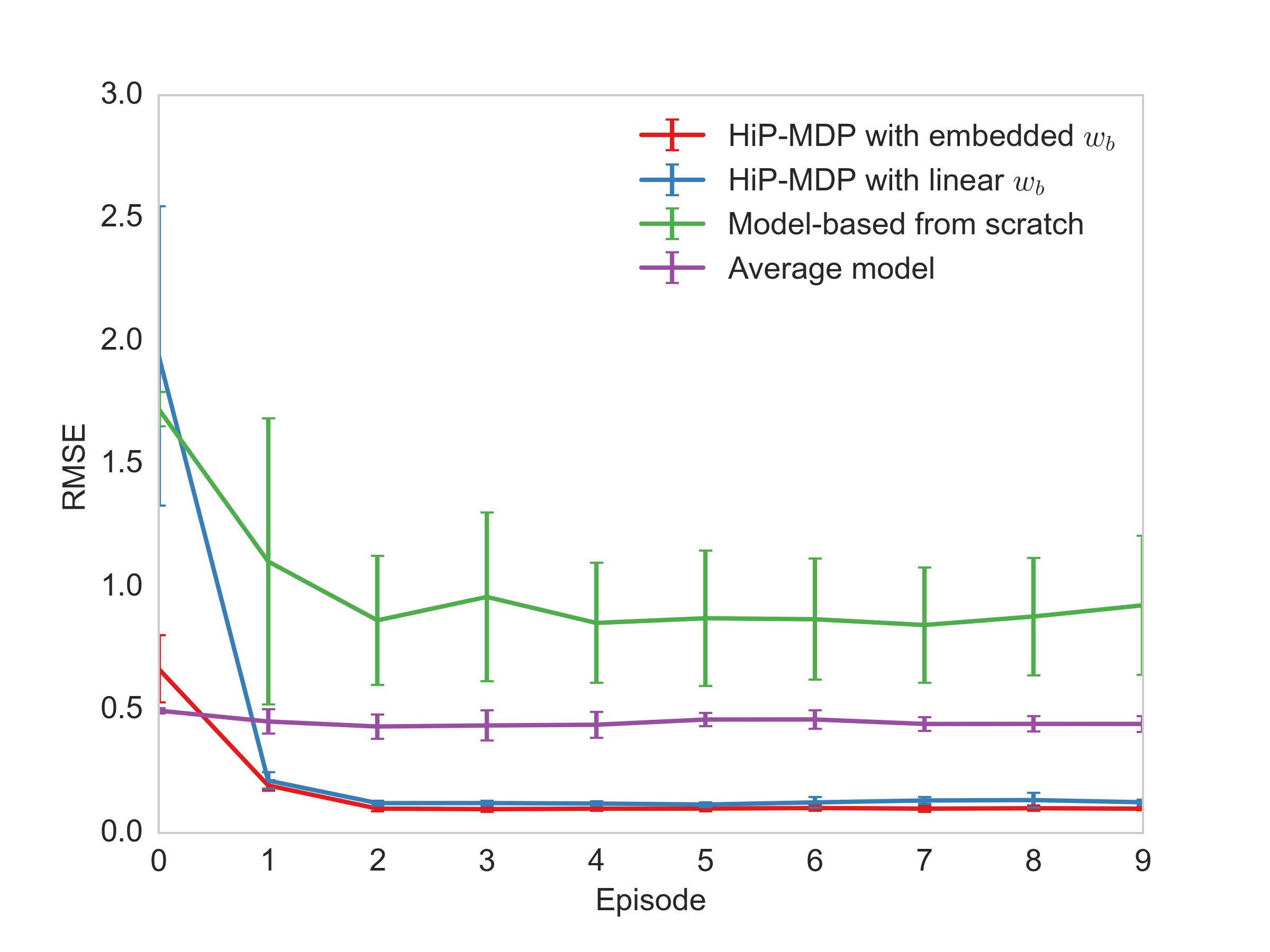}
		\caption{HIV Treatment domain  $\left[\mathcal{S}\in\mathbb{R}^6\right]$} \label{fig:hiv_erros}
	\end{subfigure}%
	\caption{Comparison of HiP-MDP transition model accuracy with the transition models trained for the baselines presented in Sec.~\ref{sec:experiments} }\label{fig:model_comparison}.
\end{figure*}

For each of these benchmarks except the BNN trained from scratch, a batch of transition data from previously observed instances was used to pre-train the BNN (and learn the latent parameters for the HiP-MDPs). Each method was then used to learn the dynamics of a new, previously unobserved instance. After the first episode in the newly encountered instance, the BNN is updated. In the two models that use a latent estimation of the environment, the latent parameters are also updated. As can be seen in Fig~\ref{fig:model_comparison}, the models using a latent parameterization improve greatly after those first network and latent parameter updates. The other two (the model from scratch and average model) also improve, but only marginally. The average model is unable to account for the different dynamics of the new instance, and the model trained from scratch does not have enough observed transition data from the new instance to construct an accurate representation of the transition dynamics.

The superior predictive performance of the two models that learn and utilize a latent estimate of the underlying dynamics of the environment reinforces the intent of the HiP-MDP as a latent variable model. That is, by estimating and employing a latent estimate of the environment, one may robustly transfer trained transition models to previously unseen instances. Further, as is shown across both domains represented in Fig.~\ref{fig:model_comparison}, the BNN with the latent parametrization embedded with the input is more reliably accurate over the duration the model interacts with a new environment. This is because the HiP-MDP with embedded latent parameters can model nonlinear interactions between the latent parameters and the state and the HiPMDP with linear latent parameters cannot. Moreover, the 2D navigation domain was constructed such that the true transition function is a nonlinear function of the latent parameter and the state. Therefore, the most accurate predictions can only be made with an approximate transition function that can model those nonlinear interactions. Hence, the 2D navigation domain demonstrates the importance of embedding the latent parameters $w_b$ with the input of the transition model.

\section{Experimental Domains}
\label{apdx:domains}

This section outlines the nonlinear dynamical systems that define the experimental domains investigated in Sec.~\ref{sec:experiments}. Here we outline the equations of motion, the hidden parameters dictating the dynamics of that motion, and the procedures used to perturb those parameters to produce subtle variations in the environmental dynamics. Other domain specific settings such as the length of an episode are also presented.

\subsection{2D Navigation Domain}
\label{apdx:dom_grid}

As was presented in Sec.~\ref{sec:model} the transition dynamics of the 2D navigation domain follow:
\begin{align*}
\Delta x &= (-1)^{\theta_b}c\big(a_x - (1-\theta_b)\beta\sqrt{(x+1.5)^2+ (y+1.5)^2}\big)\\
\Delta y &= (-1)^{\theta_b}c\big(a_y - \theta_b\beta\sqrt{(x+1.5)^2+ (y+1.5)^2}\big)\\
a_x &=
\begin{cases}
1 & a \in \{E,W\}\\
0 & \text{otherwise}
\end{cases}\\
a_y&=
\begin{cases}
1 & a \in \{N,S\}\\
0 & \text{otherwise}
\end{cases}
\end{align*}

Where $c=0.3$ and $\beta = 0.23$ are hyperparameters that restrict the agent's movement either laterally or vertically, depending on the hidden parameter $\theta_b$. In this domain, this hidden parameter is simply a binary choice $\left(\theta_b\in\{0,1\}\right)$ between the two classes of agent (``blue'' or ``red''). This force, used to counteract, or accentuate, certain actions of the agent is scaled nonlinearly by the distance the agent moves away from the center of the region of origin. 

The agent accumulates a small negative reward (-0.1) for each step taken, a large penalty (-5) if the agent hits a wall or attempts to cross into the goal region over the wrong boundary. The agent receives a substantial reward (1000) once it successfully navigates to the goal region over the correct boundary. This value was purposefully set to be large so as to encourage the agent to more rapidly enter the goal region and move against the force pushing the agent away from the goal region.

At the initialization of a new episode, the class of the agent is chosen with uniform probability and the starting state of the agent is randomly chosen to lie in the region $[-1.75,-1.25]^2$.

\subsection{Acrobot}
\label{apdx:dom_acrobot}

The acrobot domain~\citep{sutton1998rl} is dictated by the following dynamical system evolving the state parameters $s = \left[\theta_1,\theta_2,\dot{\theta}_1,\dot{\theta}_2\right]$:

\begin{align*}
\ddot{\theta}_1 &= -d_1^{-1}(d_2\ddot{\theta}_2+\phi_1) \\
\ddot{\theta}_2 &=\left(m_2l_{c2}^2 + I_2 - \frac{d_2^2}{d_1}\right)^{-1}\left(\tau + \frac{d_2}{d_1}\phi_1 - m_2l_1l_{c2}\dot{\theta}_1^2\sin\theta_2 - \phi_2\right)\\
d_1 &= m_1l_{c1}^2+m_2(l_1^2+l_{c2}^2+2l_1l_{c2}\cos\theta_2) + I_1 + I_2\\
d_2 &= m_2(l_{c2}^2+l_1l_{c2}\cos\theta_2)+I_2\\
\phi_1 &= -m_2l_1l_{c2}\dot{\theta}_2^2\sin\theta_2 - 2m_2l_1l_{c2}\dot{\theta}_2\dot{\theta}_1\sin\theta_2 + (m_1l_{c1} + m_2l_1)g\cos(\theta_1 - \pi/2) + phi_2\\
\phi_2 &= m_2l_{c2}g\cos(\theta_1 + \theta_2 - \pi/2).
\end{align*}

With reward function $R(s,a) = -0.05\left(\left[-l_1\cos(\theta_1) - l_2 \cos(\theta_1 + \theta_2)\right]- l_1\right)^2$ if the foot of the pendulum has not exceeded the goal height. If it has, then $R(s,a) = 10$ and the episode ends. The hyperparameter settings are $l_{c1}=l_{c2} = 0.5$ (lengths to center of mass of links), $I_1=I_2=1$ (moments of inertia of links) and $g = 9.8$ (gravity). The hidden parameters $\theta_b$ are the lengths and masses of the two links $(l_1,l_2,m_1,m_2)$ all set to 1 initially. In order to observe varied dynamics from this system we perturb $\theta_b$ by adding Gaussian ($\mathcal{N}(0,0.25)$) noise to each parameter independently at the initialization of a new instance. The possible state values for the angular velocities of the pendulum are constrained to $\dot{\theta}_1\in\left[-4\pi,4\pi\right]$ and $\dot{\theta}_2\in\left[-9\pi,9\pi\right]$.

At the initialization of a new episode the agent's state is initialized to $s=(0,0,0,0)$ and perturbed by some small uniformly distributed noise in each dimension. The agent is then free to apply torques to the hinge, until it raises the foot of the pendulum above the goal height or after 400 time steps.

\subsection{HIV Treatment}
\label{apdx:dom_hiv}

The dynamical system used to simulate a patient's response to HIV treatments was formulated in~\citep{adams2004dynamic}. The equations are highly nonlinear in the parameters and are used to track the evolution of six core markers used to infer a patient's overall health. These markers are, the viral load ($V$), the number of healthy and infected CD4$^{+}$ T-lymphocytes ($T_1$ and $T^{*}_1$, respectively), the number of healthy and infected macrophages ($T_2$ and $T^{*}_2$, respectively), and the number of HIV-specific cytotoxic T-cells ($E$). Thus, $s = (V,T_1,T_2,T_1^*,T_2^*,E)$. The system of equations is defined as:
\begin{align*}
\dot{T}_1 &= \lambda_1 - d_1T_1 - (1-\epsilon_1)k_1VT_1\\
\dot{T}_2 &= \lambda_2 -d_2T_2 - (1-f\epsilon_1)k_2VT_2\\
\dot{T}^*_1 &= (1-\epsilon_1)k_1VT_1-\delta T_1^*-m_1ET_2^*\\
\dot{T}^*_2 &= (1-\epsilon_2)N_T\delta(T_1^*+T_2^*)-cV-[(1-\epsilon_1)\rho_1k_1T_1 + (1-f\epsilon_1)\rho_2k_2T_2]V\\
\dot{E} &= \lambda_E + \frac{b_E(T_1^*+T_2^*)}{(T_1^*+T_2^*)+K_b}E - \frac{d_E(T_1^*+T_2^*)}{(T_1^*+T_2^*)+K_d}E-\delta_EE.\\
\end{align*}

With reward function $R(s,a) = -0.1 V - 2e^4\epsilon_1^2 - 2e^3\epsilon_2^2 + 1e^3 E$, where $\epsilon_1,\epsilon_2$ are treatment specific parameters, selected by the prescribed action.

\begin{figure}[h!]
	\centering
	\includegraphics[width=0.7\linewidth]{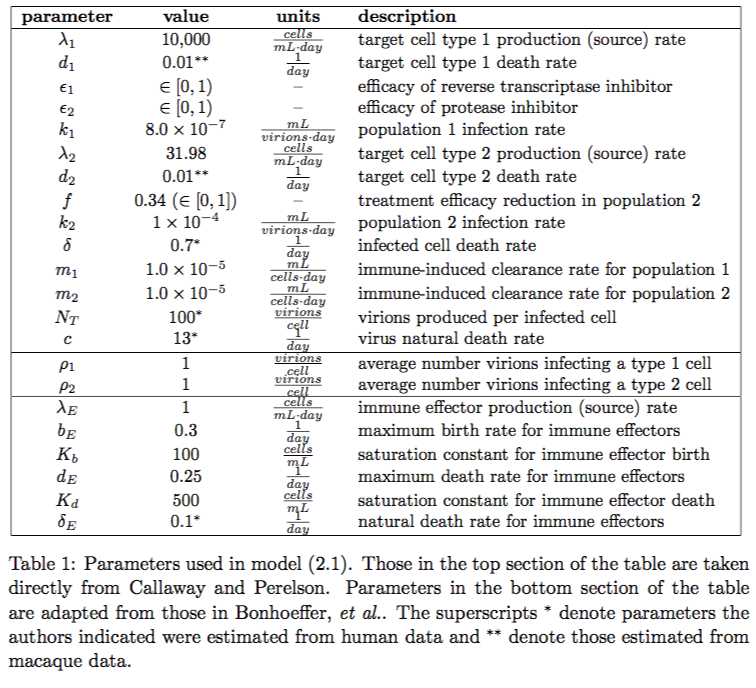}
	\caption{The hidden parameters that dictate the system dynamics of the HIV Treatment domain with their baseline values. Table courtesy of~\citet{adams2004dynamic}. }
	\label{fig:adams_params}
\end{figure}

The hidden parameters $\theta_b$ with their baseline settings~\citep{adams2004dynamic} are shown in Fig.~\ref{fig:adams_params}.

As was done with the Acrobot at the initialization of a new instance, these hidden parameters are perturbed with by some Gaussian noise ($\mathcal{N}(\theta_{b,i},0.25)$) each parameter independently. These perturbations were applied naively and at times would cause the dynamical system to lose stability or otherwise provide non-physical behavior. We filter out such instantiations of the domain and deploy the HiP-MDP on well-behaved and controllable versions of this dynamical system. 

At the initialization of a new episode the agent is started at an unhealthy steady state $$s = [163573, 5, 11945, 46, 63919, 24],$$ where the viral load and number of infected cells are much higher than the number of virus fighting T-cells. An episode is characterized by 200 time steps where, dynamically, one time step is equivalent to 5 days. At each 5 day interval, the patient's state is taken and is prescribed a treatment until the treatment period (or 1000 days) has been completed.

\section{Experiment Specifications}
\label{apdx:model_specs}
\subsection{Bayesian neural network}
\paragraph{HiP-MDP  architecture}For all domains, we model the dynamics using a feed-forward BNN. For the toy example, we used 3 fully connected hidden layers with 25 hidden units each, and for the acrobot and HIV domains, we used 2 fully connected hidden layers with 32 units each.  We used rectifier activation functions, $\phi(x)=\max(x,0)$, on each hidden layer, and the identity activation function, $\phi(x)=x$, on the output layer. For the HiP-MDP with embedded $w_b$, the input to the BNN is a vector of length $D+|A|+|w_b|$ consisting of the state $s$, a one-hot encoding of the action $a$, and the latent embedding $w_b$.  The BNN architecture for the the HiP-MDP with linear $w_b$ uses a different input layer and output layer.  The BNN input does not include the latent parameters.  Rather the BNN output, $\hat{T}^{(BNN)}(s,a)$ is a matrix of shape $ |w_b|\times D$. The next state is computed as $s' = w_b^T\hat{T}^{(BNN)}(s,a)$. In all experiments, the BNN output is the state difference $(s'-s)$ rather than the next state $s'$.

\paragraph{Hyperparameters and Training}  For all domains, we put zero mean priors on the random input noise and the network weights with variances of 1.0 and $e^{-10}$, respectively, following the procedure used by \citet{hernandez2016black}. In our experiments, we found the BNN performed best when initialized with a small prior variance on the network weights that increases over training, rather than using a large prior variance. Following \citet{hernandez2016black}, we learn the network parameters by minimizing the $\alpha$-divergence using ADAM with $\alpha=0.5$ for acrobot and the toy example and $\alpha=0.45$ for the HIV domain.  In each update to the BNN, we performed 100 epochs of ADAM, where in each epoch we sampled 160 transitions from a prioritized experience buffer and divided those transitions into mini batches of size 32. We used a learning rate of $2.5\text{\sc{e}-}{4}$ for HIV and acrobot and learning rate of $5\text{\sc{e}-}{5}$ for the toy example.

The BNN and latent parameters were learned from a batch of transition data gathered from multiple instances across 500 episodes per instance.  For the toy example, acrobot, and HIV, we use data from 2, 8, and 5 instances, respectively. For HIV, we found performance improved by standardizing the observed states to have zero mean, unit variance. 

\subsection{Latent Parameters}
For all domains, we used $|w_b|=5$ latent parameters. The latent parameters were updated using the same update procedure as for updating the BNN network parameters (except with the BNN network parameters held fixed) with a learning rate of $5\text{\sc{e}-}{4}$.

\subsection{Deep Q-Network}
To learn a policy for a new task instance, we use a Double Deep Q Network with two full connected hidden layers with 256 and 512 hidden units, respectively.  Rectifier activation functions are used for the hidden layers and the identity function is used on for the output layer. For all domains, we update the primary network weights every $N_\pi=10$ time steps using ADAM with a learning rate of $5\text{\sc{e}-}{4}$, and slowly update the target network to mirror the primary network with a rate of $\tau=0.005$.  Additionally, we clip gradients such that the L2-norm is less than 2.5. We use an $\epsilon$-greedy policy starting with $\epsilon=1.0$ and decaying $\epsilon$ after each episode (each real episode for the model-free approach and each approximated episode for the model-based approaches) with a rate of $0.995$. 

In model-based approaches, we found that the DQN learns more robust policies (both on the BNN approximate dynamics and the real environment) from training exclusively off the approximated transitions of the BNN. After training the BNN off the first episode, we train the DQN using an initial batch of $N_f=500$ approximated episodes generated using the BNN.

\subsection{Prioritized Experience Replay Buffers}
We used a TD-error-based prioritized experience replay buffer~\citep{schaul2015prioritized} to store experiences used to train the DQN. For model-based approaches, we used a separate squared-error-based prioritized buffer to store experiences used to train the BNN and learn the latent parameterization. Each prioritized buffer was large enough to store all experiences. We used a prioritization exponent of 0.2 and an importance sampling exponent of 0.1.

\section{Long run demonstration of policy learning}
\label{apdx:ex_results}
\begin{figure*}[h!]
	\begin{subfigure}[t]{0.495\textwidth}
		\centering	
		\includegraphics[width=\linewidth]{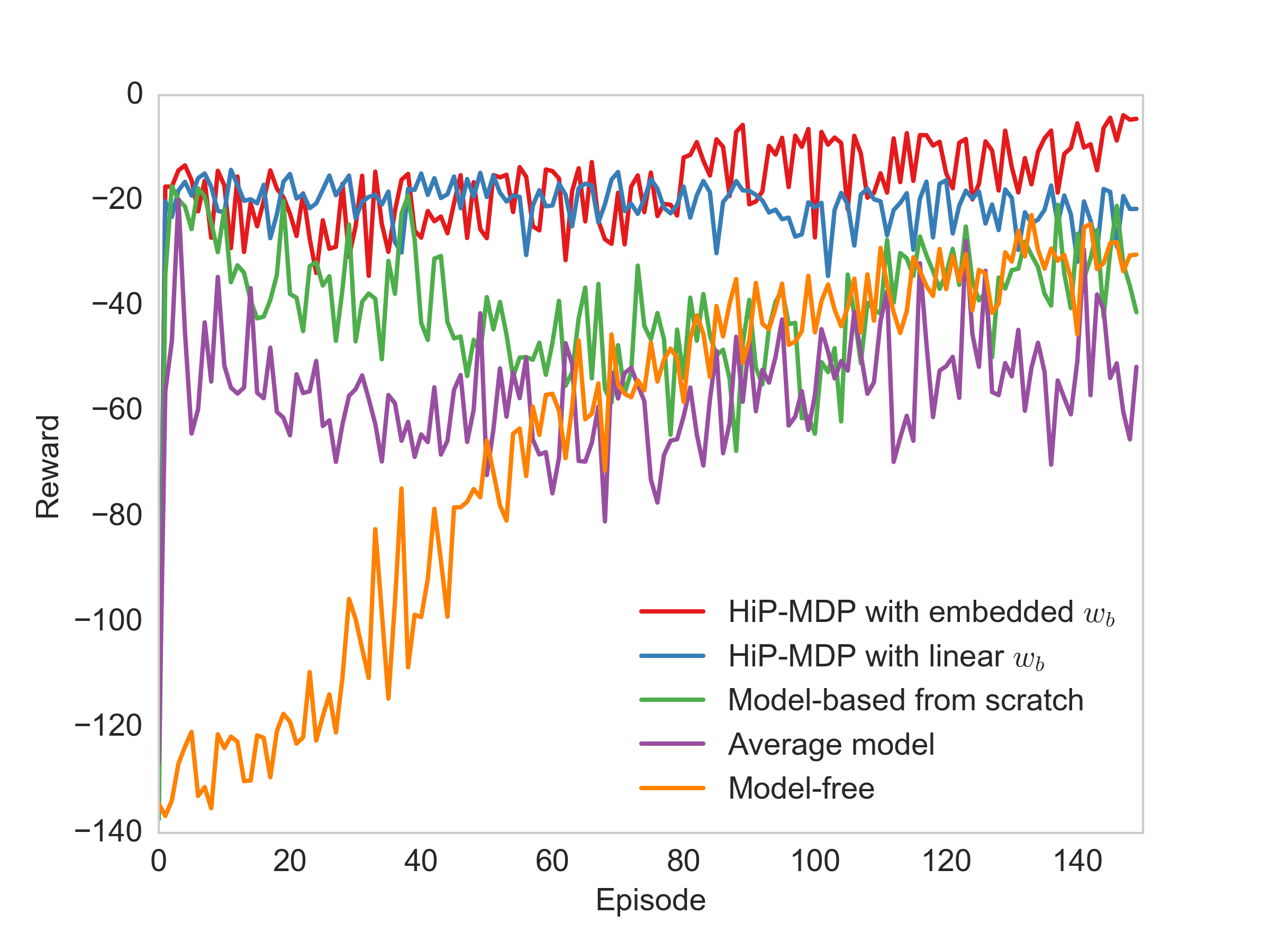}
		\caption{Acrobot} \label{fig:acrobot_500}
	\end{subfigure}%
	\begin{subfigure}[t]{0.495\textwidth}
		\centering	
		\includegraphics[width=\linewidth]{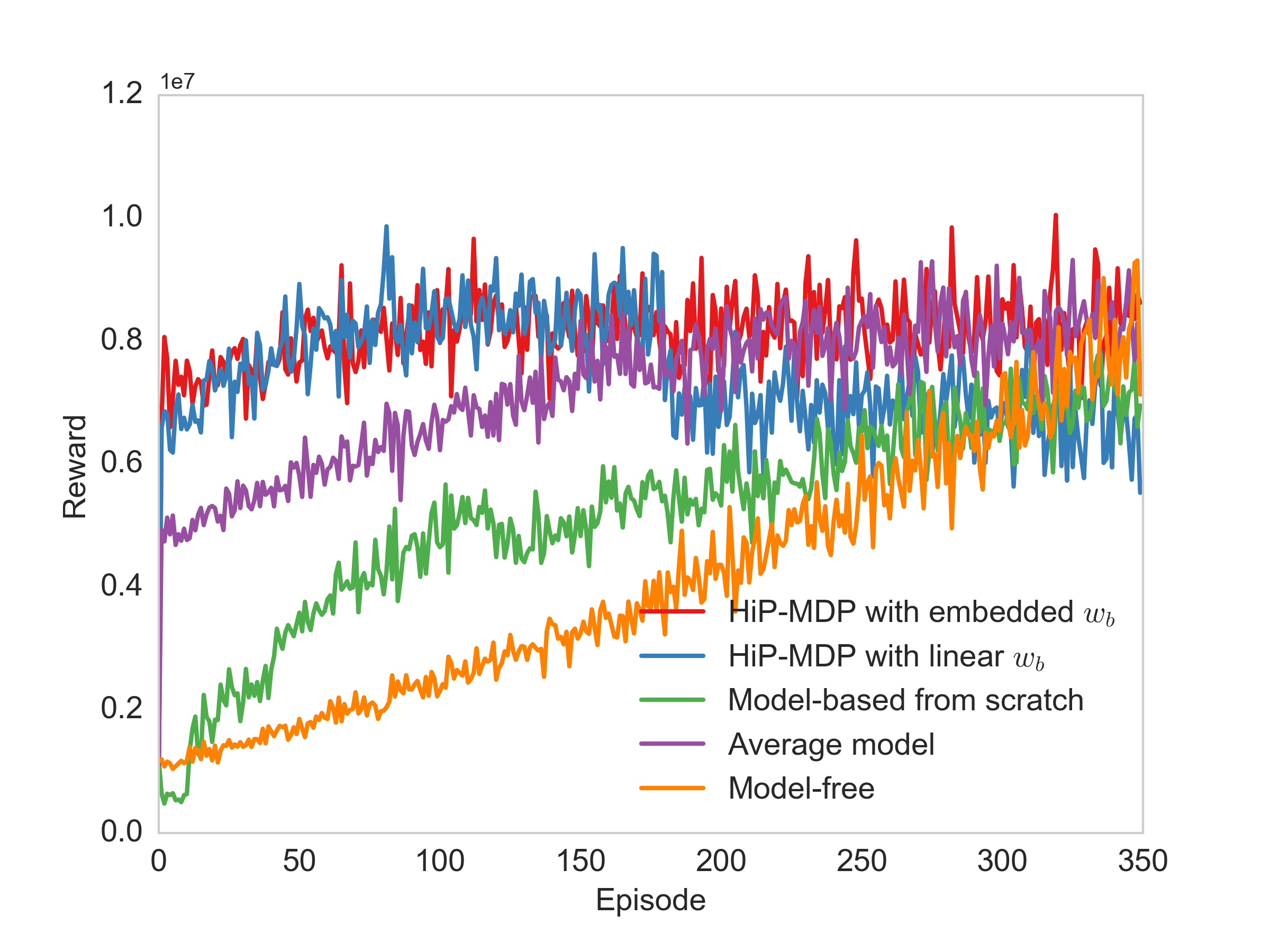}
		\caption{HIV Treatment} \label{fig:hiv_500}
	\end{subfigure}%
	\caption{A comparison of learning a policy for a new task instance $b$ using the HiP-MDP versus four benchmarks over more episodes. The mean reward for each episode over 5 runs is shown for each benchmark.  The error bars are omitted to show the results clearly.}\label{fig:policy_500}
\end{figure*}

We demonstrate that all benchmark methods used learn good control policies for new, unique instances of the acrobot domain (Figure~\ref{fig:acrobot_500}) and the HIV treatment domain (Figure~\ref{fig:hiv_500}) with a sufficient number of training episodes.
However, in terms of policy learning efficiency and proficiency, comparing the performance of the HiP-MDP with the benchmark over the first 10 episodes is instructive---as presented the Experiments section of the main paper. This format emphasizes the immediate returns of using the embedded latent parameters to transfer previously learned information when encountering a new instance of a task. 

\end{document}